\documentclass{article}
\usepackage{spconf,amsmath,graphicx}
\usepackage[dvipsnames, svgnames, x11names]{xcolor} 
\usepackage[square, comma, sort&compress, numbers]{natbib}
\usepackage{arydshln}
\usepackage{multicol}
\usepackage{epsfig}
\usepackage{epstopdf}
\usepackage{graphicx}
\usepackage{citesort}
\usepackage{amssymb}
\usepackage{amsmath}
\usepackage{color}
\usepackage{url}
\usepackage{array}
\usepackage{multirow}
\usepackage{bm}
\usepackage{tablefootnote}
\usepackage{threeparttable}
\usepackage{color}
\usepackage{colortbl}
\usepackage{mdwmath,mdwtab,amsmath}
\usepackage{autobreak}
\usepackage{amsmath,amssymb}
\usepackage{array}
\usepackage{multirow}
\usepackage{tabularx}
\usepackage{makecell}
\usepackage{graphicx}
\usepackage{threeparttable}
\usepackage{tablefootnote}
\usepackage[OT1]{fontenc}
\usepackage{float}
\newcolumntype{Y}{>{\centering\arraybackslash}X}
\usepackage[ruled,linesnumbered]{algorithm2e}
\usepackage{tabularx}
\usepackage[hang,flushmargin]{footmisc}
\usepackage{ragged2e}
\usepackage{fancyhdr}
\usepackage{bbding} 
\usepackage{indentfirst}
\usepackage{stfloats}
\usepackage{xpatch}
\usepackage{etoolbox}
\makeatletter
\usepackage{indentfirst}
\usepackage{bbding}

\usepackage[numbers,sort&compress]{natbib}
\title{Global-Local Detail Guided Transformer \\ for Sea Ice Recognition in Optical Remote Sensing Images}

\name{Zhanchao Huang*, Wenjun Hong, and Hua Su}
\address{The Key Laboratory of Spatial Data Mining and Information Sharing of Ministry of Education, \vspace{1.2mm}\\  The Academy of Digital China, Fuzhou University, Fuzhou 350108, China}

\begin{document}
\maketitle

\begin{abstract}
The recognition of sea ice is of great significance for reflecting climate change and ensuring the safety of ship navigation. Recently, many deep learning based methods have been proposed and applied to segment and recognize sea ice regions. However, the diverse scales of sea ice areas, the zigzag and fine edge contours, and the difficulty in distinguishing different types of sea ice pose challenges to existing sea ice recognition models. In this paper, a Global-Local Detail Guided Transformer (GDGT) method is proposed for sea ice recognition in optical remote sensing images. In GDGT, a global-local feature fusiont mechanism is designed to fuse global structural correlation features and local spatial detail features. Furthermore, a detail-guided decoder is developed to retain more high-resolution detail information during feature reconstruction for improving the performance of sea ice recognition. Experiments on the produced sea ice dataset demonstrated the effectiveness and advancement of GDGT.
\end{abstract}

\begin{keywords}
sea ice recognition, image segmentation, deep learning, Transformer model 
\end{keywords}

\section{Introduction}
Sea ice recognition is vital for understanding the Earth's climate system, predicting weather patterns, preserving ecosystems, ensuring safe navigation, and making informed decisions regarding resource management and conservation in the face of ongoing climate change \cite{10184008, 8320045, 10312772}. However, sea ice covers a wide area, has diverse ice categories, and sea ice areas are constantly changing with time, ocean currents, and tides, which makes it difficult to segment and recognize large-scale sea ice accurately and efficiently.

With the development of deep learning techniques, various deep learning based semantic segmentation and recognition methods have been proposed \cite{ronneberger2015u, unetformer}. For example, Kang et. al. \cite{9793723} proposed a CNN-improved encoder-decoder structure for effictive sea-ice segmentation. MeltPondNet \cite{9914571} combines Swin Transformer \cite{swin} and UNet \cite{ronneberger2015u} models for detection of melt ponds on arctic sea ice. However, the existing deep learning based sea ice segmentation and recognition methods still face the following challenges. 
On the one hand, sea ice covers a wide range and has huge scale differences. It is difficult for existing methods to accurately segment sea ice areas of different scales, especially floating ice areas with tortuous and finely divided outlines. For instance, although the UNet \cite{ronneberger2015u} structure integrates local multi-scale features, it lacks learning of global feature correlation information, resulting in low segmentation recognition accuracy of large-scale ice. On the other hand, existing methods cannot recognize different categories of sea ice in a fine-grained manner, especially in thin ice areas where the separation from sea water is not obvious. Although the Transformer structure \cite{swin} has a larger receptive field and learns non-local correlation features, the missing of spatial details leads to difficulties in segmenting fine sea ice and recognizing different categories of sea ice. In addition, the texture of the ice surface and the ice in the land area will also interfere with sea ice segmentation and recognition.

In this regard, a Global-Local Detail Guided Transformer (GDGT) method is proposed for sea ice segmentation and recognition in optical remote sensing images. The contributions of this work are summarized as follows:

1) A global-local feature fusiont (GLFF) mechanism is designed in decoder to fuse global structural correlation features and local spatial detail features for the more accurate segmentation of multi-scale sea ice areas. 

2) A detail-guided decoder (DGD) is developed to use discrete wavelet features to guide learnable filtering, allowing the decoder to retain more high-resolution detail information during feature reconstruction and improve the performance of fine sea ice recognition.

\section{The Proposed GDGT Model}

\begin{figure*}[!t]
	\centering
	\epsfig{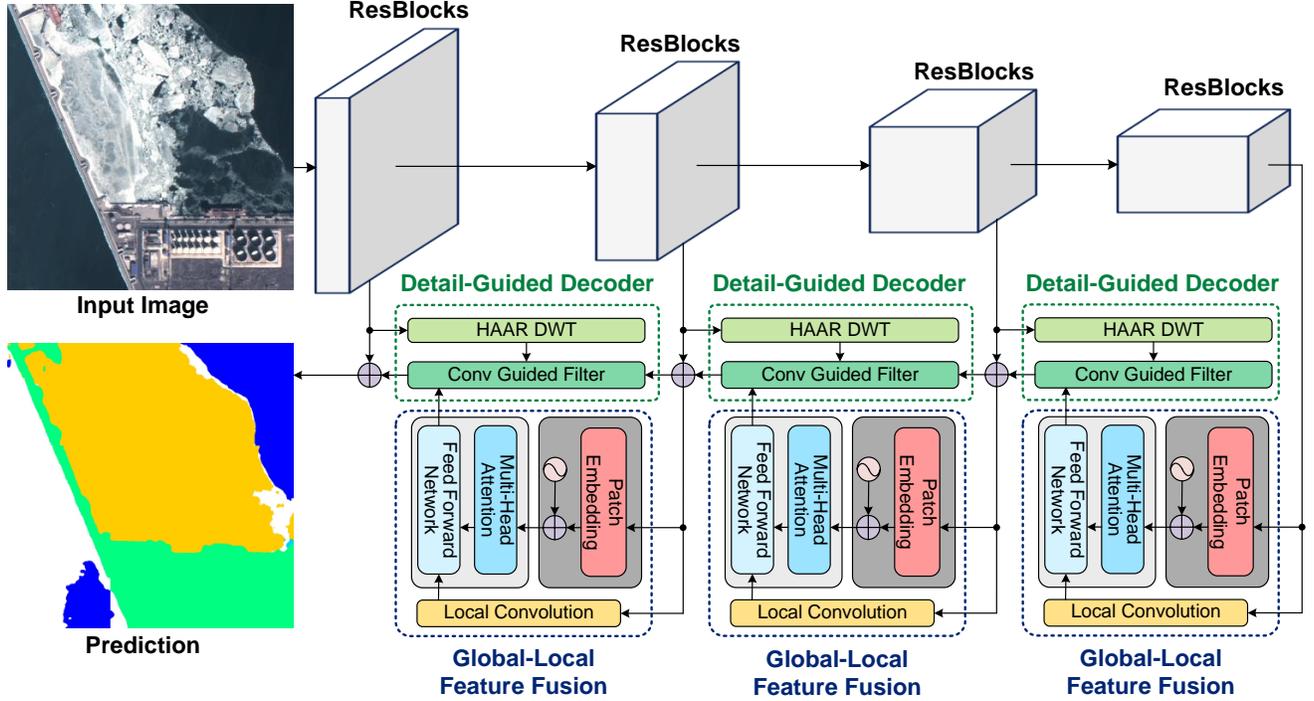}
	\caption{The framework of the proposed GDGT for seaice segmentation and recognition.}\label{fig:1}
\end{figure*}

The proposed global-local detail guided Transformer framework is shown in Fig.~\ref{fig:1}. The model uses the U-shaped structure of UNet, in which the down-sampling encoder adopts the ResNet structure and the up-sampling decoder adopts the Transformer structure. Feature interaction occurs between the encoder and the decoder through latent connections. To enable the decoder to learn global structural features and local detail features at the same time, the Transformer blocks of the decoder are improved into GLFF modules for more comprehensive multi-scale feature fusion. Furthermore, to enable the decoder to retain more high-resolution detail information during feature reconstruction, a DGD module is designed to improve the latent connection between the encoder and decoder. High-resolution detailed features from the encoding stage are retained through discrete wavelet transform (DWT) to guide feature upsampling in the decoding stage. The GLFF module and DGD module will be explained in detail below.

\begin{figure}[!t]
	\centering
	\epsfig{width=0.48\textwidth,file=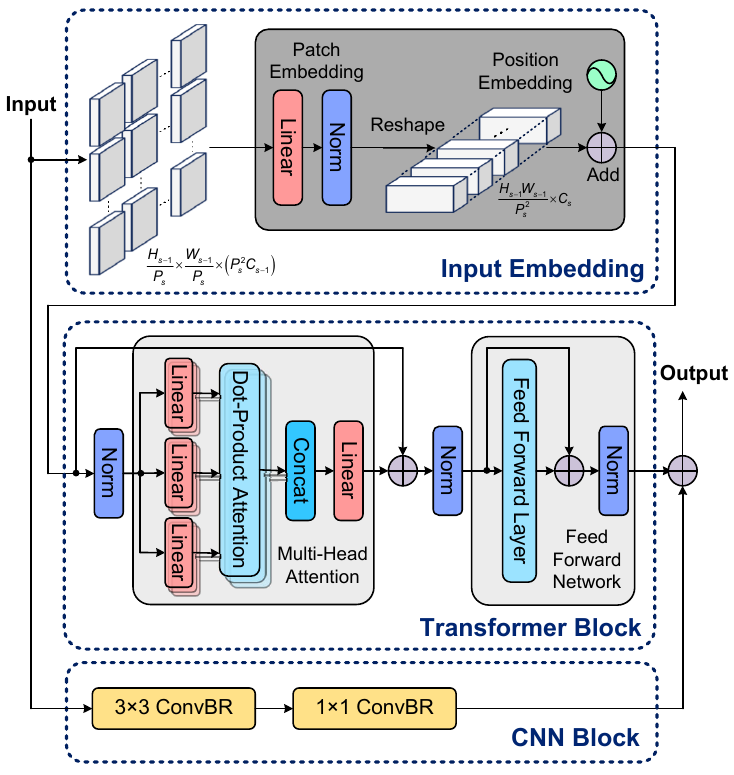}
	\caption{The Global-Local Feature Fusion (GLFF) Module.}\label{fig:2}
\end{figure}
\begin{figure}[t]
	\centering
	\epsfig{width=0.49\textwidth,file=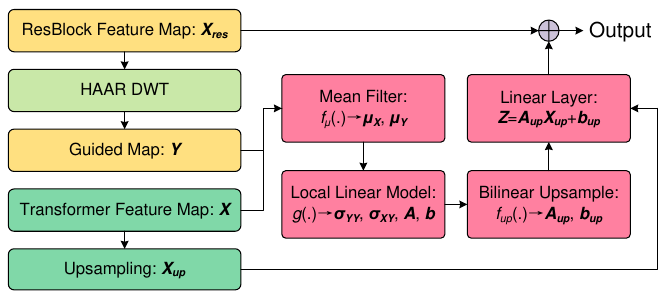}
	\caption{The Detail-Guided Decoder (DGD) Module. }\label{fig:3}
\end{figure}

\subsection{Global-Local Feature Fusion}

The features extracted by the CNN convolutional layer contain rich spatial detail information. However, the features lack non-local long-distance dependence, which will lead to inaccurate segmentation of large-scale sea ice area contours. Transformer learns global feature correlation information through self-attention but misses local spatial details, which will make it difficult for the model to identify some sea ice areas that are fragmented and have tortuous edge contours. In this regard, a global-local feature fusion module GLFF shown in Fig.~\ref{fig:2} is designed in the decoder of GDGT. 

First, the upsampled encoder features are divided into several embedding patches to save the calculation of self-attention\cite{swin}. Then, after passing through the linear layer, three sets of feature vectors are obtained as query, key and value input to the multi-head attention block. Then, the non-local correlation features of the sea ice area are obtained through a feed-forward operation. At the same time, the decoder features are fed into an additional convolution branch to extract local detail features. Finally, a set of adaptive weights are learned to perform a weighted fusion of global and local features to obtain global-local features for better prediction.

\subsection{Detail-Guided Decoder}

During the down-sampling encoding and up-sampling decoding processes of the model, the edge details of the object are easily missed. However, fine segmentation of sea ice regions relies on these detailed texture and contour features. Therefore, the DGD module as shown in Fig.~\ref{fig:3} is designed. 

First, the shallow semantic features with higher resolution are decomposed by HAAR wavelet to obtain high-frequency detailed features. Denote the shallow semantic features from a residual block of the encoder as $\boldsymbol{X}_{res}$, which have

\begin{equation}
	{\boldsymbol{LL},\boldsymbol{LH},\boldsymbol{HL},\boldsymbol{HH}} = dwt(\boldsymbol{X}_{res}),
	\label{eq:0}
\end{equation}
where $dwt(.)$ represents the discrete HAAR wavelet transform. The low-frequency component $\boldsymbol{LL}$, horizontal high-frequency component $\boldsymbol{LH}$, vertical high-frequency component $\boldsymbol{HL}$, and diagonal high-frequency component $\boldsymbol{HH}$ obtained by HAAR wavelet decomposition, \begin{equation}
	\small
	\begin{array}{l}
		\! \left\{\begin{array}{l}
			\!	L{L_{i,j}} \!= \!{X_{2i - 1,2j - 1}} \!+ \!{X_{2i - 1,2j}} \!+ \!{X_{2i,2j - 1}} \!+ \!{X_{2i,2j}}\\
			\!	L{H_{i,j}} \!= \!- {X_{2i - 1,2j - 1}} \!- \!{X_{2i - 1,2j}} \!+ \!{X_{2i,2j - 1}} \!+ \!{X_{2i,2j}}\\
			\!	H{L_{i,j}} \!= \! - {X_{2i - 1,2j - 1}} \!+ \!{X_{2i - 1,2j}} \!- \!{X_{2i,2j - 1}} \!+ \!{X_{2i,2j}}\\
			\!	H{H_{i,j}} \!= \!{X_{2i - 1,2j - 1}} \!- \!{X_{2i - 1,2j}} \!- \!{X_{2i,2j - 1}} \!+ \!{X_{2i,2j}}
		\end{array} \right.
	\end{array},
	\label{eq:1}
\end{equation}
where $L{L_{i,j}}$, $L{H_{i,j}}$, $H{L_{i,j}}$ and $H{H_{i,j}}$ represent the elements of $\boldsymbol{LL}$, $\boldsymbol{LH}$, $\boldsymbol{HL}$ and $\boldsymbol{HH}$ at $(i,j)$, respectively. 

Then, through the learnable guided filtering \cite{he2012guided, wu2018fast} mechanism, detailed features are retained and enhanced in the decoder, improving the model's segmentation accuracy for sea ice areas (especially fine sea ice areas and thin ice areas). Denote the guided feature map containing rich detailed features of sea-ice images as
\begin{equation}
	\boldsymbol{Y} = Conv\left( {\boldsymbol{LL},\boldsymbol{LH},\boldsymbol{HL},\boldsymbol{HH}} \right),
	\label{eq:2}
\end{equation}
where $Conv(.)$ denotes the $1 \times 1$ convolution, is generated. 

\begin{equation}
\boldsymbol{\mu_X} = {f_\mu }\left( \boldsymbol{X} \right), \ \boldsymbol{\mu_Y} = {f_\mu }\left( \boldsymbol{Y} \right),
	\label{eq:3}
\end{equation}
where ${f_\mu }$ denotes a learnable mean filter constructed through convolutional layers. 

\begin{equation}	
	\boldsymbol{\sigma _{XY}} = g\left( {\boldsymbol{X} \cdot \boldsymbol{Y}} \right), \ \boldsymbol{\sigma _Y} = g\left( {\boldsymbol{Y} \cdot \boldsymbol{Y}} \right),
	\label{eq:4}
\end{equation}
where ${g }$ denotes a local linear operation, 
\begin{equation}	
	g\left( {\boldsymbol{X} \cdot \boldsymbol{Y}} \right) = {f_\mu }\left( {\boldsymbol{X} \cdot \boldsymbol{Y}} \right) - \boldsymbol{\mu_X} \cdot \boldsymbol{\mu_Y},
	\label{eq:5}
\end{equation}

\begin{equation}	
	\boldsymbol{A} = \frac{{{\boldsymbol{\sigma _{XY}}}}}{{{\boldsymbol{\sigma _Y}} + \boldsymbol{\varepsilon} }}, \ \boldsymbol{b} = {\boldsymbol{\mu _Y}} - \boldsymbol{A} \cdot \boldsymbol{\mu _X}.
	\label{eq:6}
\end{equation}
where $\boldsymbol{\varepsilon}$ denotes the learnable regularization term. Thus, the output of the DGD module is
\begin{equation}	
	\boldsymbol{Z} = \alpha \cdot up(\boldsymbol{A}) \cdot up(\boldsymbol{X}) + up(\boldsymbol{b}) + \beta \cdot \boldsymbol{X_{res}},
	\label{eq:7}
\end{equation}
where $up(.)$ represents the up-sampling operation, $\alpha$ and $\beta$ denote learnable weights.

\section{Experiments and Discussions}

\subsection{Experimental Conditions}
      
1) Experimental Details. All experiments are performed on a computer with an RTX4090 GPU. The maximum epoch of training is set to 12, the the initial learning rate is set to $6 \times 10 ^{-4}$, and the batch size is set to 64.

2) Dataset.
We collect and produce a sea-ice segmentation and recognition dataset based on optical remote sensing images. The images of sea ice come from the GF-2 satellite. In order to make the images cover multi-scale areas, the images are scaled by the ratio of x0.25, x0.50, x1.00, x1.50. Due to the huge size of the remote sensing images, they are cropped into sub-images of 800 × 800 pixels (overlapping 200 pixels). These images have been carefully annotated at the pixel level. There are 5 categories of objects to be recognized, including "Sea", "Thin-Ice", "Thick-Ice", ''Land' and "Pool-Ice". When input to the deep learning models for training and testing, these images are scaled to 512 × 512 pixels. The training dataset and testing dataset contain 10918 and 1842 pairs of images and ground truth, respectively.

\begin{table}[b]
	\renewcommand\arraystretch{1.8}
	\setlength{\tabcolsep}{0.5mm}{
		\caption{\label{table:1}
			{Ablation experiments of the proposed GDGT}}
		\resizebox{0.48\textwidth}{!}{
			\begin{tabular}{c|ccc|c|c|c|c}
				\hline\hline
				Method & Decoder & GLFF & DGD & mIoU (\%) & F1 (\%) & OA (\%) & FWIoU (\%) \\ 
				\hline		
				UNet \cite{ronneberger2015u} & CNN & $\times$ & $\times$ & 80.21 & 88.05 & 91.69 & 85.97 \\
				UNetFormer\cite{unetformer} & Trans. & $\times$ & $\times$ & 86.88 & 92.55 & 95.06 & 91.06 \\
				+GLFF & Trans. & $\checkmark$ & $\times$ & 87.52 & 92.93 & 95.41 & 91.53\\
				+GLFF+DGD & Trans. & $\checkmark$ & w/o. DWT & 88.12 & 93.33 & 95.54 & 91.81\\
    \hline
				\textbf{GDGT} & Trans. & $\checkmark$ & w. DWT & \textbf{88.57} & \textbf{93.62} & \textbf{95.76} & \textbf{92.14}  \\	
				
				\hline\hline
		\end{tabular}}
	}
\end{table}
\begin{table*}[t]
	\renewcommand\arraystretch{1.2}
	\setlength{\tabcolsep}{0.5mm}{
		\caption{\label{table:2}
			{Performance comparison of different semantic segmentation methods}}
		\resizebox{\textwidth}{!}{\setlength{\tabcolsep}{1mm}{
				\begin{threeparttable}
					\begin{tabular}{c|ccccc|c|c|c|c|c|c}
						\hline\hline
						Methods & Sea (\%) & Thin-Ice (\%) & Thick-Ice (\%) & Land (\%)& Pool Ice (\%) & mIoU (\%) & F1 (\%) & OA (\%) & FWIoU (\%) & Para. (M) & Size (MB)\\
						\hline
						ABCNet \cite{abcnet} & 82.65 & 87.27 & 59.19 & 98.34 & 77.31 & 80.95 & 88.89 & 93.97 & 89.27 & 14.00 & 55.85\\
						A$^2$FPN \cite{a2fpn} & 87.42 & \textbf{90.75} & 68.08 & \textbf{98.71} & 86.62 & 86.32 & 92.33 & 95.75 & 92.15 & 12.20 & 48.64\\					
						MANet \cite{manet} & 81.90 & 84.98 & 55.27 & 98.31 & 78.20 & 82.99 & 89.95 & 93.05 & 88.15 & 35.90 & 143.44 \\
						BANet \cite{banet} & 80.89 & 87.48 & 62.14 & 98.49 & 77.04 & 84.21 & 90.88 & 94.03 & 89.45 & 12.70 & 50.92\\
						
                        DC-Swin \cite{dcswin} & 83.95 & 88.43 & 63.56 & 98.68 & 85.60 & 86.44 & 92.27 & 94.79 & 90.59 & 66.9 & 267.80 \\
						
						UNetFormer \cite{unetformer} & 86.41 & 89.31 & 64.66 & 98.51 & 83.43 & 86.88 & 92.55 & 95.06 & 91.06 & \textbf{11.70} & \textbf{46.91} \\ 
						\hline
						\textbf{GDGT}  & \textbf{86.96} & 90.73 & \textbf{68.89} & 98.62 & \textbf{86.90} &  \textbf{88.57} & \textbf{93.62} & \textbf{95.76} & \textbf{92.14} & 11.80 & 47.15\\											
						\hline\hline
					\end{tabular}
	\end{threeparttable}}}}
\end{table*}

\begin{figure}[!t]
	\centering
	\epsfig{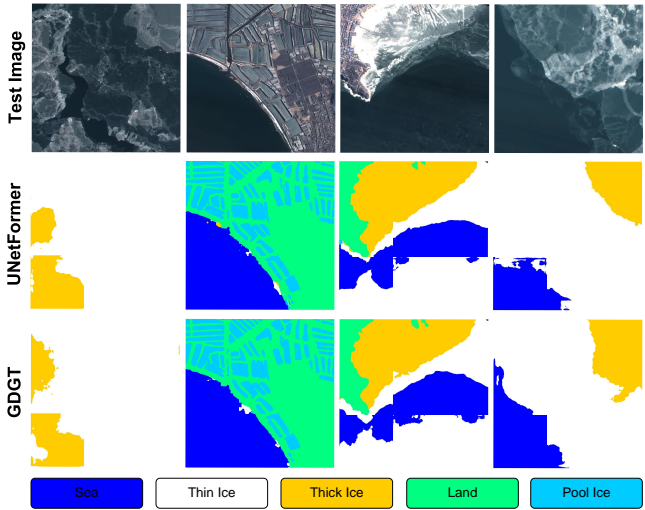}
	\caption{Visualization of comparative experimental results of predicting image patches.}\label{fig:4}
\end{figure}
\begin{figure}[!t]
	\centering
	\epsfig{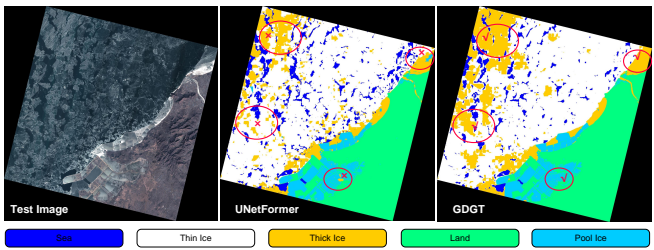}
	\caption{Visualization of prediction results for the entire remote sensing image.}\label{fig:5}
\end{figure}

\subsection{Experiments and Discussions}
Ablation results of each component of the proposed GDGT on the sea ice dataset are listed in Table~\ref{table:1}. When GLFF is used to fuse the local detail features extracted by CNN and the global structure correlation features extracted by Transformer in the decoding stage, the performance of sea ice segmentation and recognition is significantly improved compared to UNet \cite{ronneberger2015u} and UNetFormer \cite{unetformer}. This shows that the comprehensive use of high-resolution local detail features and low-correlation global structural features in the decoding stage can help improve the accuracy of sea ice segmentation and recognition. Furthermore, after introducing the designed DGD, the detailed features of the sea ice edge are retained and enhanced in the decoder upsampling process, especially when using HAAR wavelet features as the guidance map, which further improves the performance. Compared with UNet and UNetFormer, the mIoU of the proposed GDGT is improved by 7.91\% and 1.69\% respectively. Fig.~\ref{fig:4} shows the sea ice segmentation and recognition results of UNetFormer \cite{unetformer} and GDGT on image patches, verifying that the proposed GDGT can more accurately identify objects at the edge of the sea ice area.

The comparison experiments with the existing segmentation methods are listed in Table~\ref{table:2}. Several state-of-the-art (SOTA) CNN-based methods, including ABCNet \cite{abcnet}, A$^2$FPN \cite{a2fpn}, MANet \cite{manet} and Transformer-based methods, including BANet \cite{banet}, DC-Swin \cite{dcswin}, UNetFormer \cite{unetformer} are used as controls. Experiments show that the performance of the proposed GDGT exceeds that of existing methods in the four metircs of mIoU, F1-score, OA and FWIoU. Fig.~\ref{fig:5} shows the segmentation and recognition results of UNetFormer \cite{unetformer} and the proposed GDGT of the entire remote sensing image. It demonstrates that the proposed GDGT performs better in sea ice area contour segmentation and fine sea ice recognition. 

\section{Conclusions}
In this paper, a global-local detail guided Transformer method for sea ice segmentation and recognition has been proposed. In the proposed GDGT, the GLFF module combines the local detail features extracted by CNN blocks and the global structural features extracted by Transformer blocks, enhancing the multi-scale feature expression ability of the model. Morever, the developed DGD enables the feature decoding process to retain the detailed information of the edge areas of sea ice and improve the accuracy of semantic segmentation. Experiments on the produced sea-ice dataset have demonstrated that the proposed GDGT achieves the advancement performance on sea ice segmentation and recognition in optical remote sensing images.

\vfill\pagebreak
\section{Reference}
\vspace{-3em}
\bibliographystyle{IEEE.bst}
\bibliography{GDGT.bib}

\begin{thebibliography}{10}

\bibitem{10184008}
Mohammed Shokr and Mohammed Dabboor,
\newblock ``Polarimetric sar applications of sea ice: A review,''
\newblock {\em IEEE Journal of Selected Topics in Applied Earth Observations
  and Remote Sensing}, vol. 16, pp. 6627--6641, 2023.

\bibitem{8320045}
Weikai Tan, Jonathan Li, Linlin Xu, and Michael~A. Chapman,
\newblock ``Semiautomated segmentation of sentinel-1 sar imagery for mapping
  sea ice in labrador coast,''
\newblock {\em IEEE Journal of Selected Topics in Applied Earth Observations
  and Remote Sensing}, vol. 11, no. 5, pp. 1419--1432, 2018.

\bibitem{10312772}
Rafael Pires~de Lima and Morteza Karimzadeh,
\newblock ``Model ensemble with dropout for uncertainty estimation in sea ice
  segmentation using sentinel-1 sar,''
\newblock {\em IEEE Transactions on Geoscience and Remote Sensing}, vol. 61,
  pp. 1--15, 2023.

\bibitem{ronneberger2015u}
Olaf Ronneberger, Philipp Fischer, and Thomas Brox,
\newblock ``U-net: Convolutional networks for biomedical image segmentation,''
\newblock in {\em Medical Image Computing and Computer-Assisted
  Intervention--MICCAI 2015: 18th International Conference, Munich, Germany,
  October 5-9, 2015, Proceedings, Part III 18}. Springer, 2015, pp. 234--241.

\bibitem{unetformer}
Libo Wang, Rui Li, Ce~Zhang, Shenghui Fang, Chenxi Duan, Xiaoliang Meng, and
  Peter~M Atkinson,
\newblock ``Unetformer: A unet-like transformer for efficient semantic
  segmentation of remote sensing urban scene imagery,''
\newblock {\em ISPRS Journal of Photogrammetry and Remote Sensing}, vol. 190,
  pp. 196--214, 2022.

\bibitem{9793723}
Jian Kang, Fengyu Tong, Xiang Ding, Sijiang Li, Ruoxin Zhu, Yan Huang, Yusheng
  Xu, and Ruben Fernandez-Beltran,
\newblock ``Decoding the partial pretrained networks for sea-ice segmentation
  of 2021 gaofen challenge,''
\newblock {\em IEEE Journal of Selected Topics in Applied Earth Observations
  and Remote Sensing}, vol. 15, pp. 4521--4530, 2022.

\bibitem{9914571}
Ivan Sudakow, Vijayan~K. Asari, Ruixu Liu, and Denis Demchev,
\newblock ``Meltpondnet: A swin transformer u-net for detection of melt ponds
  on arctic sea ice,''
\newblock {\em IEEE Journal of Selected Topics in Applied Earth Observations
  and Remote Sensing}, vol. 15, pp. 8776--8784, 2022.

\bibitem{swin}
Ze~Liu, Yutong Lin, Yue Cao, Han Hu, Yixuan Wei, Zheng Zhang, Stephen Lin, and
  Baining Guo,
\newblock ``Swin transformer: Hierarchical vision transformer using shifted
  windows,''
\newblock in {\em Proceedings of the IEEE/CVF international conference on
  computer vision}, 2021, pp. 10012--10022.

\bibitem{he2012guided}
Kaiming He, Jian Sun, and Xiaoou Tang,
\newblock ``Guided image filtering,''
\newblock {\em IEEE transactions on pattern analysis and machine intelligence},
  vol. 35, no. 6, pp. 1397--1409, 2012.

\bibitem{wu2018fast}
Huikai Wu, Shuai Zheng, Junge Zhang, and Kaiqi Huang,
\newblock ``Fast end-to-end trainable guided filter,''
\newblock in {\em Proceedings of the IEEE Conference on Computer Vision and
  Pattern Recognition}, 2018, pp. 1838--1847.

\bibitem{abcnet}
Rui Li, Shunyi Zheng, Ce~Zhang, Chenxi Duan, Libo Wang, and Peter~M. Atkinson,
\newblock ``Abcnet: Attentive bilateral contextual network for efficient
  semantic segmentation of fine-resolution remotely sensed imagery,''
\newblock {\em ISPRS Journal of Photogrammetry and Remote Sensing}, vol. 181,
  pp. 84--98, 2021.

\bibitem{a2fpn}
Rui Li, Libo Wang, Ce~Zhang, Chenxi Duan, and Shunyi Zheng,
\newblock ``A2-fpn for semantic segmentation of fine-resolution remotely sensed
  images,''
\newblock {\em International Journal of Remote Sensing}, vol. 43, no. 3, pp.
  1131--1155, 2022.

\bibitem{manet}
Rui Li, Shunyi Zheng, Ce~Zhang, Chenxi Duan, Jianlin Su, Libo Wang, and
  Peter~M. Atkinson,
\newblock ``Multiattention network for semantic segmentation of fine-resolution
  remote sensing images,''
\newblock {\em IEEE Transactions on Geoscience and Remote Sensing}, vol. 60,
  pp. 1--13, 2022.

\bibitem{banet}
Libo Wang, Rui Li, Dongzhi Wang, Chenxi Duan, Teng Wang, and Xiaoliang Meng,
\newblock ``Transformer meets convolution: A bilateral awareness network for
  semantic segmentation of very fine resolution urban scene images,''
\newblock {\em Remote Sensing}, vol. 13, no. 16, 2021.

\bibitem{dcswin}
Libo Wang, Rui Li, Chenxi Duan, Ce~Zhang, Xiaoliang Meng, and Shenghui Fang,
\newblock ``A novel transformer based semantic segmentation scheme for
  fine-resolution remote sensing images,''
\newblock {\em IEEE Geoscience and Remote Sensing Letters}, vol. 19, pp. 1--5,
  2022.

\end{thebibliography}

\end{document}